\documentclass[10pt,twocolumn,letterpaper]{article}

\usepackage{wacv}

\usepackage{amsmath,amssymb,amsfonts,amsthm}
\usepackage{bm}
\usepackage{booktabs}
\usepackage{multirow}
\usepackage{array}
\usepackage{enumitem}
\usepackage{graphicx}
\usepackage{tikz}
\usepackage{pgfplots}
\usepackage{microtype}
\usepackage[table]{xcolor}
\usepackage{pifont}
\usetikzlibrary{arrows.meta,calc,positioning}
\usepgfplotslibrary{groupplots}
\pgfplotsset{compat=1.18}

\DeclareMathOperator*{\argmax}{arg\,max}
\DeclareMathOperator{\tr}{tr}

\newcommand{\ours}{ShrinkSel}
\newcommand{\qpr}{\textnormal{\textsc{QPR}}}

\usepackage{placeins}

\definecolor{wacvblue}{rgb}{0.21,0.49,0.74}
\usepackage[pagebackref,breaklinks,colorlinks,allcolors=wacvblue]{hyperref}

\title{Diagnosing Prototype Estimation in Few-Shot Class-Incremental Learning: Training-Free Geometry Selection and Unlabeled-Query Prototype Rectification}

\author{Md Tanvir Hossain Saikat\textsuperscript{1}\quad
Md Akib Haider\textsuperscript{1}\thanks{Equal contribution}\quad
Md. Tahmid Islam Tomal\textsuperscript{2,*}\\
Md Hasanul Kabir\textsuperscript{1}\\
\textsuperscript{1}Islamic University of Technology, Bangladesh\\
\textsuperscript{2}Bangladesh University of Engineering and Technology, Bangladesh\\
{\small\ttfamily \{tanvirhossain3, akibhaider, hasanul\}@iut-dhaka.edu}\\
{\small\ttfamily tahmid.tomal@icloud.com}}

\begin{document}
\maketitle
\raggedbottom

\newcommand{\NumDiagnosisPairs}{10}
\newcommand{\NumDiagnosisPasses}{10}
\newcommand{\ShrinkWinsCosine}{10/10}
\newcommand{\ShrinkWinsTEEN}{10/10}
\newcommand{\ShrinkWinsRanPAC}{9/10}
\newcommand{\ShrinkWinsSharedLDA}{9/10}
\newcommand{\BaseOnlyDiagnosisPasses}{0/3}

\newcommand{\NumCrossFamilies}{2}
\newcommand{\NumCrossDiagnosisRows}{6}
\newcommand{\NumCrossDiagnosisPasses}{6}
\newcommand{\NumCrossShrinkLDAWins}{6}
\newcommand{\NumCrossShrinkLDACIWins}{6}
\newcommand{\NumCrossShrinkCosineWins}{5}
\newcommand{\NumCrossShrinkRanPACWins}{6}
\newcommand{\CrossShrinkLDAMinGain}{0.50}
\newcommand{\CrossShrinkLDAMaxGain}{1.52}

\begin{abstract}
	Freezing a visual encoder prevents representation drift, but a novel class
	must still be represented using only a few labeled examples. We study how
	much this uncertainty in class means contributes to the accuracy gap between
	a five-shot cosine classifier and a complete-data linear probe. On identical
	frozen features, we first replace novel means with means from the complete
	training split, keeping base prototypes and cosine scoring fixed. We then
	fit the linear probe. Across sixteen strong encoder--dataset pairs, mean
	replacement accounts for more than half of this ordered gap; a weaker
	base-only encoder does not show the same pattern. Motivated by this finding,
	\ours{} selects prototype radius and covariance shrinkage using base-class
	episodes, with regularized shared-covariance LDA as a fallback. In the
	sixteen-pair evaluation, it improves over LDA on fifteen pairs by
	$0.04$--$1.52$ percentage points and selects LDA on Aircraft. Query Prototype
	Rectification (\qpr{}) additionally uses an unlabeled cohort to update novel
	prototype directions while preserving their radii and the selected metric.
	On the three standard benchmarks with CLIP ViT-B/16, it adds
	$0.37$--$0.67$ points over cosine NCM and $1.21$--$2.03$ points over \ours{}
	in session-average accuracy. Gains remain positive
	when each query is excluded from its own update. However, novel-class gains
	can come at the expense of base accuracy, and the benefit diminishes with
	more labeled supports or very small query cohorts.
\end{abstract}

\section{Introduction}
\label{sec:intro}

Few-shot class-incremental learning (FSCIL) adds new classes from a small
labeled support set while retaining the ability to recognize previously
learned classes \cite{tao2020fscil,zhou2024fscilsurvey}. Adapting to these few examples risks overfitting to new
classes and forgetting previously learned ones. Many methods modify
the representation or classifier to balance these objectives
\cite{zhang2021cec,zhou2022fact,yang2023ncfscil,song2023savc}. Freezing the
encoder prevents representation drift. Accuracy can still fall, however, as the classifier must distinguish more classes.

A prototype classifier illustrates the remaining difficulty. Base-class
means are estimated from many examples, whereas each novel-class mean may
use only five. An inaccurate novel mean can move a decision boundary even
when the underlying features are useful. We therefore ask how much accuracy can be recovered by estimating novel-class means more accurately while leaving the classifier unchanged.

We examine this question through two comparisons on identical frozen
features. First, we replace five-shot novel means with means computed from
the complete training split, keeping base prototypes and cosine scoring
fixed. Second, we fit a linear probe to the same complete training features.
Across sixteen settings using strong CLIP, MAE-ft, and supervised ResNet features, mean replacement recovers more than half of the gap in this ordered comparison. Results across support counts and individual classes are consistent with estimation error from limited samples. The weaker base-only ResNet does not follow the same pattern, so this finding is limited to the representations and comparison studied here.

\begin{figure*}[!t]
	\centering
	\includegraphics[width=\textwidth]{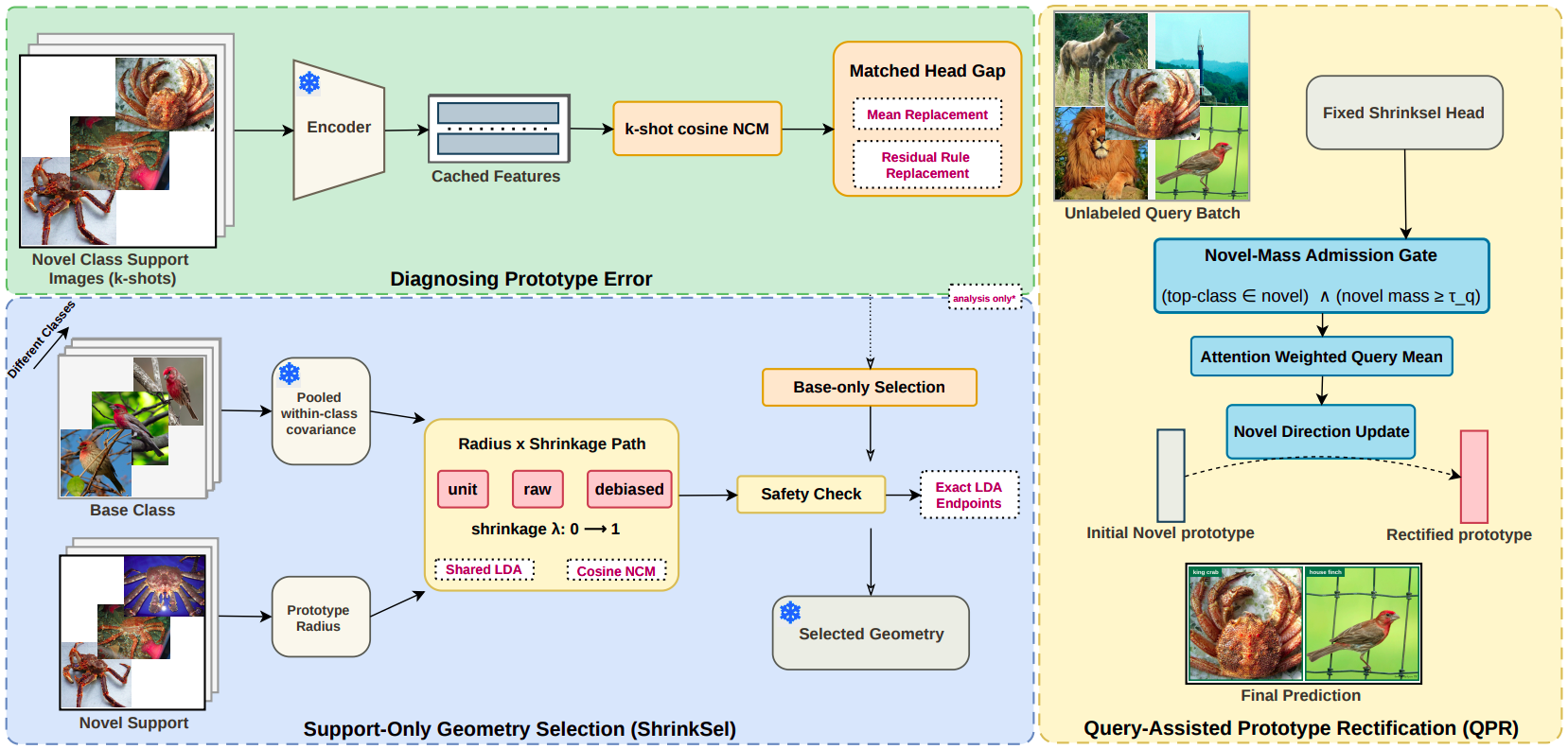}
	\caption{Overview of the analysis and the two prediction methods, all using
		frozen image features. The retrospective analysis replaces five-shot novel
		means with means from the complete training split, then fits a linear probe.
		\ours{} uses base classes to select a support-only classifier. \qpr{} adds an
		unlabeled query cohort to refine novel prototype directions. Only the analysis
		uses complete novel training labels; neither prediction method uses query
		labels or known query-class proportions.}
	\label{fig:overview}
\end{figure*}

These results motivate two methods that use different sources of data
(Fig.~\ref{fig:overview}). \ours{} estimates a support-only head from
labeled examples and base statistics. When an unlabeled cohort is
available, \qpr{} refines novel prototype directions while preserving
base prototypes, novel radii, and the metric. We evaluate \qpr{} with
both \ours{} and cosine NCM to separate its
contribution from the choice of support-only head.
Neither prediction method uses text or known query-class proportions.

Our contributions are:
\begin{itemize}[leftmargin=1.2em]
	\item A controlled study of novel-mean replacement followed by classifier replacement on frozen features, with uncertainty estimates and analyses across support counts and individual classes.
	\item \ours{}, a support-only procedure that selects prototype radius
	and covariance shrinkage using base-class episodes, with an LDA fallback.
	\item \qpr{}, a method for refining novel prototype directions using unlabeled queries. We evaluate it with two support-only heads, exclude each query from its own update, vary cohort size, and examine the trade-off between base and novel accuracy.
\end{itemize}

\section{Related Work}
\label{sec:related}

\paragraph{FSCIL with fixed or adapted representations.}
FSCIL methods often change the representation or classifier to retain previously learned classes while learning new ones from few examples. CEC~\cite{zhang2021cec} evolves the classifier, FACT~\cite{zhou2022fact} reserves
space for future classes, and NC-FSCIL~\cite{yang2023ncfscil}, OrCo~\cite{ahmed2024orco}, and ASP~\cite{liu2024asp} introduce geometric constraints or prompt-based
structure.
With strong pretrained features, Wu et al.~\cite{wu2022strongptm} adapt a novel-class
branch and combine its scores with those of the base classifier. FSA~\cite{panos2023fsa} adapts the
representation only in the first session, while RanPAC~\cite{mcdonnell2023ranpac} and
RanDumb~\cite{prabhu2024randumb} use analytic heads over fixed feature expansions. We freeze the encoder in
every session and isolate uncertainty in the resulting head.
\paragraph{Support-only estimation.}
Cosine NCM~\cite{mensink2013ncm} and prototypical networks~\cite{snell2017protonet}
represent a class by its support mean. TEEN~\cite{wang2023teen} transfers base-class
structure to scarce classes. Distribution Calibration~\cite{yang2021freelunch}
synthesizes features, whereas higher-order calibration~\cite{goswami2024higherorder}
combines novel and semantically related base covariances after base-session adapter
tuning. EPT~\cite{jiang2026ept} learns prototype offsets.
FeCAM~\cite{fecam2023} and HyCal~\cite{lee2026hycal} use covariance-aware distances,
whereas ULA~\cite{woo2026ula} adjusts base--novel logits. These approaches differ in their use of adaptation and synthetic data. We keep the features fixed and compute the head in closed form. Using class-disjoint base episodes, we select prototype radius together with covariance shrinkage~\cite{ledoit2004shrinkage}. The candidate family includes cosine NCM and shared LDA as exact endpoints.

\paragraph{Different modality budgets.}
PriViLege~\cite{park2024privilege}, FDR~\cite{xue2025fdr}, BiMC~\cite{bimc2025},
TIM++~\cite{li2026timpp}, and Transductive CLIP~\cite{martin2024transclip} draw on text,
prompts, language-derived initialization, or vision--text probability features.
These methods provide useful multimodal contexts, but their inputs differ from
those available to our frozen image-only heads. We therefore report them separately
from the primary rankings.

\paragraph{Transductive inference.}
Transductive few-shot methods use unlabeled queries to refine prototypes or predictions. TIM~\cite{boudiaf2020tim} maximizes mutual information,
LaplacianShot~\cite{ziko2020laplacianshot} encourages smooth predictions on a query graph, PT-MAP~\cite{hu2021ptmap} uses optimal transport, and ProtoLP~\cite{zhu2023protolp}
alternates label propagation with prototype refinement. These methods are usually evaluated in balanced episodes containing only
a few novel classes. Also their performance can depend on the
class proportions assumed for the query set~\cite{veilleux2021realistic}.
Prototype Rectification~\cite{liu2020protrect} already uses unlabeled queries
to refine prototypes through label propagation and feature shifting. We study query-assisted refinement in all-seen FSCIL, where both base and novel classes are evaluated. Our \qpr{} keeps base prototypes unchanged and adjusts only
eligible novel prototype directions. The support-selected radii and metric remain fixed, and \qpr{} does not need to know the query proportions.

\paragraph{Diagnosis.}
Approximation-risk analysis~\cite{wang2024approxrisk} examines transfer, margins, and
base--novel balance in FSCIL. We instead study the accuracy recovered by replacing
novel-class means on fixed features before changing the classifier.

\section{Diagnosing and Correcting Prototype Error}
\label{sec:method}

\subsection{Setting and information regimes}

Session $0$ provides labeled base classes $Y_0$; session $t>0$ adds a disjoint set
$Y_t$ with $K$ labeled examples per class. Prediction after session $t$ covers
$\mathcal C_t=\bigcup_{j=0}^{t}Y_j$. A frozen image encoder produces
\begin{equation}
 z(x)=f_{\rm img}(x)/\lVert f_{\rm img}(x)\rVert_2\in\mathbb R^D .
 \label{eq:feature}
\end{equation}
Our CLIP pipeline calls only the image tower. We call $Y_t$, $t>0$,
\emph{head-novel}: these identities are new to the downstream classifier, but their
absence from encoder pretraining cannot generally be verified.

A \emph{support-only} rule may use $z(x)$, labeled supports, and base statistics. A
\emph{query-assisted} rule may also use an unlabeled cohort $\mathcal U_t$ containing
only classes in $\mathcal C_t$ and is therefore batch-dependent. Complete labeled
novel data are restricted to the retrospective diagnosis. ``Training-free'' below
means no gradient-based parameter fitting; closed-form base statistics and base-only
selection remain allowed. Known query-class proportions are not evaluated as a
deployable method.

\subsection{Ordered matched counterfactual path}
\label{sec:accounting}

For support set $S_c^K$ and full training split $T_c$, define
\begin{equation}
 \textstyle
 \hat\mu_c^K=\operatorname{norm}\!\bigl(\sum_{z\in S_c^K}z\bigr),\,
 \hat\mu_c^{\rm full}=\operatorname{norm}\!\bigl(\sum_{z\in T_c}z\bigr),
 \label{eq:means}
\end{equation}
where $\operatorname{norm}(v)=v/\lVert v\rVert_2$ for $v\ne0$. The complete-split
mean is a finite labeled reference, not a population oracle. At the final session,
$A_n^K$ and $A_n^{\rm full}$ are class-balanced novel accuracies from cosine NCM with
base means fixed and only novel means changed. Let $A_n^{\rm LP}$ be the accuracy of a
fixed $\ell_2$-regularized multinomial probe trained on the same complete features for
all seen classes. The ordered path obeys
\begin{equation}
 \underbrace{A_n^{\rm LP}-A_n^K}_{\text{matched head gap}}=
 \underbrace{A_n^{\rm full}-A_n^K}_{\text{mean replacement}}+
 \underbrace{A_n^{\rm LP}-A_n^{\rm full}}_{\text{residual rule change}} .
 \label{eq:accounting}
\end{equation}
This is an exact accounting identity, not a unique causal decomposition. We report the
ordered share $r=(A_n^{\rm full}-A_n^K)/(A_n^{\rm LP}-A_n^K)$ only when a crossed
support-by-class bootstrap interval excludes a zero denominator. Probe, bootstrap,
repair, and classwise definitions are in the supplement.

\subsection{Finite-shot motivation}
\label{sec:finite-shot}

For an unnormalized mean $\hat\mu_c=\mu_c+\epsilon$ with
$\mathbb E[\epsilon]=0$, $\operatorname{Cov}(\epsilon)=\Sigma/K$, and $\epsilon$
independent of the query feature $x$, let $G$ be a fixed positive-definite metric and
$d_G(a,b)=(a-b)^\top G(a-b)$. Then
\begin{equation}
 \mathbb E\!\left[d_G(x,\hat\mu_c)-d_G(x,\mu_c)\right]
 =\tr(G\Sigma)/K.
 \label{eq:finite-penalty}
\end{equation}
Equation~\eqref{eq:finite-penalty} describes the expected distance penalty for the
correct class when its mean is estimated from a finite sample. It does not provide
an accuracy bound or account for prototype normalization, competing classes, or
estimation of $G$. The metric affects both estimation noise and class separation,
so the equation motivates selecting this trade-off but does not determine a
shrinkage value.

\subsection{Support-only geometry selection}
\label{sec:shrinksel}

\ours{} estimates the pooled within-class covariance from complete base features,
\begin{equation}
\begin{aligned}
 \bar z_c^B&=|T_c|^{-1}\sum_{z\in T_c}z,\quad
 \delta=\max\{\tr(\bar\Sigma_B)/D,10^{-12}\},\\
 \bar\Sigma_B&=\frac{1}{M}\sum_{c\in Y_0}\sum_{z\in T_c}
 (z-\bar z_c^B)(z-\bar z_c^B)^\top,
\end{aligned}
 \label{eq:base-covariance}
\end{equation}
where $T_c$ is the complete base training set and
$M=\sum_c(|T_c|-1)$. This covariance is frozen after the base session.

For $n$ i.i.d. unit support features with sample mean $\bar z_c$ and population mean
$\mu_c$, the raw resultant length is upward biased because
\begin{equation}
 \mathbb E\lVert\bar z_c\rVert_2^2=
 \frac1n+\frac{n-1}{n}\lVert\mu_c\rVert_2^2 .
 \label{eq:resultant-bias}
\end{equation}
For $n\ge2$, we therefore use
\begin{equation}
 \hat r_c=\sqrt{\bigl[
 (n\lVert\bar z_c\rVert_2^2-n^{-1}\sum\nolimits_i\lVert z_i\rVert_2^2)/(n-1)
 \bigr]_+}.
 \label{eq:resultant-correction}
\end{equation}
For $n=1$, we use $\hat r_c=\lVert\bar z_c\rVert_2$.
Here $[a]_+=\max(a,0)$. Clipping ensures a valid radius but can reintroduce
finite-sample bias. The prototype family is unit
$u_c=\operatorname{norm}(\bar z_c)$, raw $\bar z_c$, or corrected $\hat r_cu_c$.
Each is crossed with
\begin{equation}
\begin{aligned}
 G_\lambda&=[(1-\lambda)\bar\Sigma_B+(\lambda+\gamma)\delta I]^{-1},
 \quad \gamma=1,\\
 s_c(z)&=-\tfrac12(z-p_c)^\top G_\lambda(z-p_c),\\
 \lambda&\in\{0,.1,.25,.5,.75,.9,1\}.
\end{aligned}
 \label{eq:metric-score}
\end{equation}
Unit prototypes at $\lambda=1$ reproduce cosine NCM, while raw prototypes at
$\lambda=0$ reproduce ridge-regularized shared-covariance LDA with equal class
priors.

\paragraph{Base-only selection.}
Let $B=|Y_0|$, let each of $T$ increments contain $m$ classes, and define the
class-count weights
\begin{equation}
 w_n=\frac{1}{T+1}\sum_{t=0}^{T}\frac{tm}{B+tm},\qquad w_b=1-w_n .
 \label{eq:selector-weights}
\end{equation}
For candidate $h=(p,\lambda)$, each pseudo-increment measures anchor and pseudo-novel
accuracy, $A_b(h)$ and $A_n(h)$, and scores
$J(h)=w_bA_b(h)+w_nA_n(h)$. We first divide the base-class identities into construction, selection, and safety partitions. For \ours{}, construction and selection identities
are pooled and deterministically repartitioned into four class-disjoint pseudo-novel
folds; safety identities never enter selection. Across 200 episodes, a candidate is
feasible only if it does not trail raw $\lambda=0$ LDA in any fold. The feasible
candidate with highest pooled $J$ is checked on safety; a regression beyond $0.10$
point deploys the exact LDA endpoint. After selection, we refit the chosen geometry using the permitted base features and freeze it. The supplement specifies the split sizes, seeds, episode construction, tie-breaking rules, and cumulative-count stress test.

\subsection{Query-assisted prototype rectification}
\label{sec:qpr}

Query Prototype Rectification (\qpr{}) starts from a fixed support-only classifier.
Our main pipeline (Fig. ~\ref{fig:overview}) uses \ours{}. Table~\ref{tab:transductive-benchmark} also evaluates
the same update with cosine NCM, which uses unit-norm prototypes and cosine scores.
Initial scores define
\begin{equation}
\begin{aligned}
 \pi(c\mid z)&=\operatorname{softmax}_{c\in\mathcal C_t}(\kappa s_c(z)),\\
 m_n(z)&=\sum_{c\in\mathcal C_t\setminus Y_0}\pi(c\mid z).
\end{aligned}
 \label{eq:qpr-posterior}
\end{equation}
A novel class $c$ admits
\begin{equation}
\begin{gathered}
 \mathcal G_c=\{z\in\mathcal U_t:
 \argmax_{d\in\mathcal C_t}\pi(d\mid z)\notin Y_0,\\
 m_n(z)\ge\tau_q,\quad
 c=\argmax_{d\in\mathcal C_t\setminus Y_0}\pi(d\mid z)\}.
\end{gathered}
 \label{eq:qpr-gate}
\end{equation}
Class $c$ retains its original prototype when fewer than $m_{\min}$ queries are
admitted, that is, when $|\mathcal G_c|<m_{\min}$. Otherwise, define
$a_c(z)=\pi(c\mid z)/\sum_{u\in\mathcal G_c}\pi(c\mid u)$ and perform
\begin{equation}
\begin{gathered}
 \tilde\mu_c=\operatorname{norm}\!\bigl({\textstyle\sum_{z\in\mathcal G_c}a_c(z)z}\bigr),
 \quad p'_c=\lVert p_c\rVert_2v'_c,\\
 v'_c=\operatorname{norm}((1-\alpha)\operatorname{norm}(p_c)
 +\alpha\tilde\mu_c).
\end{gathered}
 \label{eq:qpr-update}
\end{equation}
Base prototypes, non-admitted novel prototypes, and $G_\lambda$ remain fixed; all
queries are then rescored with the same support-only rule: Eq.~\eqref{eq:metric-score}
for \ours{}, or cosine scoring for the cosine NCM comparison. The update is one-shot,
restarts from the support-only head each session, and assumes neither labels nor class
proportions. Ordinary QPR uses the same cohort to construct and evaluate the update.
Exact self-exclusion subtracts query $z_i$ from its admitted aggregate when classifying
$z_i$, while retaining all other cohort queries. The four controls are selected on identity-disjoint base pseudo-query episodes spanning
balanced and deterministic-imbalanced cohorts, with held-out combined and anchor
safety gates; failure deploys the exact no-update case $\alpha=0$. The 96-point grid,
gate thresholds, calibration episodes, tie-breaking, vectorized self-exclusion, and
cumulative-competition calibration are detailed in the supplement.

\section{Experiments}
\label{sec:exp}

\subsection{Experimental design}

\paragraph{Data and encoders.}
We use BiMC's~\cite{bimc2025} five-shot class orders and manifests.
CUB-200~\cite{wah2011cub} uses $100+10\times10$ classes;
CIFAR-100~\cite{krizhevsky2009cifar} and miniImageNet~\cite{vinyals2016matching}
use $60+8\times5$.
Aircraft~\cite{maji2013aircraft} and DTD~\cite{cimpoi2014describing} provide
additional domain checks. Results average twenty matched support draws,
with nested supports for $K\in\{1,2,5,10,20\}$.

The primary matrix uses frozen OpenAI CLIP~\cite{radford2021clip} ViT-B/16,
ViT-B/32, and RN50 image features. MAE-ft ViT-B/16~\cite{he2022mae} and supervised
ImageNet ResNet-50~\cite{he2016resnet} provide cross-family checks.
A ResNet-18 trained only on benchmark base classes provides a weaker control.

\paragraph{Comparators.}
Support-only comparisons include NCM~\cite{mensink2013ncm}, shared and diagonal LDA,
TEEN~\cite{wang2023teen}, ULA~\cite{woo2026ula}, FeCAM~\cite{fecam2023},
RanPAC~\cite{mcdonnell2023ranpac}, RanDumb~\cite{prabhu2024randumb}, and an image-only
reduction of HyCal~\cite{lee2026hycal}. Query-assisted comparisons include
TIM-GD~\cite{boudiaf2020tim}, LaplacianShot~\cite{ziko2020laplacianshot},
ProtoLP~\cite{zhu2023protolp}, and uniform-marginal MAP-OT$_u$~\cite{hu2021ptmap}.
Matched comparisons share features, supports, sessions, and queries.
Table~\ref{tab:transductive-benchmark} evaluates every query-assisted method
with both cosine NCM and \ours{} as support-only references. The added
\ours{} variants use its prototypes and a common whitening transform of
supports and queries, normalized by a base-only scalar. They retain each
method's update, including full LaplacianShot's CL2N preprocessing.
Existing base-selected settings are held fixed across heads. The supplement
describes the original native-pipeline comparisons; Table~\ref{tab:transductive-benchmark}
extends them with the additional head comparisons.

\begin{figure*}[!t]
	\centering

\begin{tikzpicture}
\begin{groupplot}[group style={group size=3 by 1,horizontal sep=0.82cm},
width=0.305\textwidth,height=4.7cm,xmin=-2,xmax=42,xtick={0,10,20,30,40},
ymin=-0.45,ymax=2.45,ytick={0,1,2},
xmajorgrids=true,grid style={black!10},
tick label style={font=\scriptsize},title style={font=\small},
legend style={font=\scriptsize,draw=none,legend columns=2,
/tikz/every even column/.append style={column sep=0.3cm}}]
\nextgroupplot[title={CUB-200},yticklabels={RN50,ViT-B/32,ViT-B/16},legend to name=decomposition-legend]
\draw[densely dashed,black!55,line width=0.45pt] (axis cs:0,-0.42)--(axis cs:0,2.42);
\draw[blue!70,line width=1.05pt] (axis cs:0,2.10)--(axis cs:18.451,2.10);
\draw[orange!88!black,line width=1.05pt,-{Latex[length=2.2pt,width=1.7pt]},shorten <=2.2pt,shorten >=2.2pt] (axis cs:18.451,2.10)--(axis cs:16.108,1.90);
\draw[blue!85!black,line width=0.55pt] (axis cs:16.533,2.10)--(axis cs:20.349,2.10);
\draw[blue!85!black,line width=0.55pt] (axis cs:16.533,2.035)--(axis cs:16.533,2.165);
\draw[blue!85!black,line width=0.55pt] (axis cs:20.349,2.035)--(axis cs:20.349,2.165);
\draw[orange!88!black,line width=0.55pt] (axis cs:13.682,1.90)--(axis cs:18.518,1.90);
\draw[orange!88!black,line width=0.55pt] (axis cs:13.682,1.835)--(axis cs:13.682,1.965);
\draw[orange!88!black,line width=0.55pt] (axis cs:18.518,1.835)--(axis cs:18.518,1.965);
\draw[blue!70,line width=1.05pt] (axis cs:0,1.10)--(axis cs:21.446,1.10);
\draw[orange!88!black,line width=1.05pt,-{Latex[length=2.2pt,width=1.7pt]},shorten <=2.2pt,shorten >=2.2pt] (axis cs:21.446,1.10)--(axis cs:19.517,0.90);
\draw[blue!85!black,line width=0.55pt] (axis cs:19.463,1.10)--(axis cs:23.416,1.10);
\draw[blue!85!black,line width=0.55pt] (axis cs:19.463,1.035)--(axis cs:19.463,1.165);
\draw[blue!85!black,line width=0.55pt] (axis cs:23.416,1.035)--(axis cs:23.416,1.165);
\draw[orange!88!black,line width=0.55pt] (axis cs:17.064,0.90)--(axis cs:21.916,0.90);
\draw[orange!88!black,line width=0.55pt] (axis cs:17.064,0.835)--(axis cs:17.064,0.965);
\draw[orange!88!black,line width=0.55pt] (axis cs:21.916,0.835)--(axis cs:21.916,0.965);
\draw[blue!70,line width=1.05pt] (axis cs:0,0.10)--(axis cs:21.446,0.10);
\draw[orange!88!black,line width=1.05pt,-{Latex[length=2.2pt,width=1.7pt]},shorten <=2.2pt,shorten >=2.2pt] (axis cs:21.446,0.10)--(axis cs:20.170,-0.10);
\draw[blue!85!black,line width=0.55pt] (axis cs:19.524,0.10)--(axis cs:23.372,0.10);
\draw[blue!85!black,line width=0.55pt] (axis cs:19.524,0.035)--(axis cs:19.524,0.165);
\draw[blue!85!black,line width=0.55pt] (axis cs:23.372,0.035)--(axis cs:23.372,0.165);
\draw[orange!88!black,line width=0.55pt] (axis cs:17.749,-0.10)--(axis cs:22.651,-0.10);
\draw[orange!88!black,line width=0.55pt] (axis cs:17.749,-0.165)--(axis cs:17.749,-0.035);
\draw[orange!88!black,line width=0.55pt] (axis cs:22.651,-0.165)--(axis cs:22.651,-0.035);
\addplot+[forget plot,only marks,mark=o,mark size=1.65pt,draw=black!65,mark options={solid,fill=white}] coordinates {(0,2.10) (0,1.10) (0,0.10)};
\addplot+[only marks,mark=*,mark size=2.0pt,blue!85!black] coordinates {(18.451,2.10) (21.446,1.10) (21.446,0.10)};
\addlegendentry{Complete-mean cosine}
\addplot+[only marks,mark=square*,mark size=2.0pt,orange!88!black] coordinates {(16.108,1.90) (19.517,0.90) (20.170,-0.10)};
\addlegendentry{Complete-data linear probe}
\nextgroupplot[title={CIFAR-100},yticklabels={,,}]
\draw[densely dashed,black!55,line width=0.45pt] (axis cs:0,-0.42)--(axis cs:0,2.42);
\draw[blue!70,line width=1.05pt] (axis cs:0,2.10)--(axis cs:23.294,2.10);
\draw[orange!88!black,line width=1.05pt,-{Latex[length=2.2pt,width=1.7pt]},shorten <=2.2pt,shorten >=2.2pt] (axis cs:23.294,2.10)--(axis cs:32.444,1.90);
\draw[blue!85!black,line width=0.55pt] (axis cs:20.244,2.10)--(axis cs:26.408,2.10);
\draw[blue!85!black,line width=0.55pt] (axis cs:20.244,2.035)--(axis cs:20.244,2.165);
\draw[blue!85!black,line width=0.55pt] (axis cs:26.408,2.035)--(axis cs:26.408,2.165);
\draw[orange!88!black,line width=0.55pt] (axis cs:28.717,1.90)--(axis cs:36.181,1.90);
\draw[orange!88!black,line width=0.55pt] (axis cs:28.717,1.835)--(axis cs:28.717,1.965);
\draw[orange!88!black,line width=0.55pt] (axis cs:36.181,1.835)--(axis cs:36.181,1.965);
\draw[blue!70,line width=1.05pt] (axis cs:0,1.10)--(axis cs:24.235,1.10);
\draw[orange!88!black,line width=1.05pt,-{Latex[length=2.2pt,width=1.7pt]},shorten <=2.2pt,shorten >=2.2pt] (axis cs:24.235,1.10)--(axis cs:34.735,0.90);
\draw[blue!85!black,line width=0.55pt] (axis cs:20.945,1.10)--(axis cs:27.554,1.10);
\draw[blue!85!black,line width=0.55pt] (axis cs:20.945,1.035)--(axis cs:20.945,1.165);
\draw[blue!85!black,line width=0.55pt] (axis cs:27.554,1.035)--(axis cs:27.554,1.165);
\draw[orange!88!black,line width=0.55pt] (axis cs:30.635,0.90)--(axis cs:38.845,0.90);
\draw[orange!88!black,line width=0.55pt] (axis cs:30.635,0.835)--(axis cs:30.635,0.965);
\draw[orange!88!black,line width=0.55pt] (axis cs:38.845,0.835)--(axis cs:38.845,0.965);
\draw[blue!70,line width=1.05pt] (axis cs:0,0.10)--(axis cs:20.980,0.10);
\draw[orange!88!black,line width=1.05pt,-{Latex[length=2.2pt,width=1.7pt]},shorten <=2.2pt,shorten >=2.2pt] (axis cs:20.980,0.10)--(axis cs:36.680,-0.10);
\draw[blue!85!black,line width=0.55pt] (axis cs:18.291,0.10)--(axis cs:23.732,0.10);
\draw[blue!85!black,line width=0.55pt] (axis cs:18.291,0.035)--(axis cs:18.291,0.165);
\draw[blue!85!black,line width=0.55pt] (axis cs:23.732,0.035)--(axis cs:23.732,0.165);
\draw[orange!88!black,line width=0.55pt] (axis cs:33.324,-0.10)--(axis cs:40.073,-0.10);
\draw[orange!88!black,line width=0.55pt] (axis cs:33.324,-0.165)--(axis cs:33.324,-0.035);
\draw[orange!88!black,line width=0.55pt] (axis cs:40.073,-0.165)--(axis cs:40.073,-0.035);
\addplot+[forget plot,only marks,mark=o,mark size=1.65pt,draw=black!65,mark options={solid,fill=white}] coordinates {(0,2.10) (0,1.10) (0,0.10)};
\addplot+[only marks,mark=*,mark size=2.0pt,blue!85!black] coordinates {(23.294,2.10) (24.235,1.10) (20.980,0.10)};
\addplot+[only marks,mark=square*,mark size=2.0pt,orange!88!black] coordinates {(32.444,1.90) (34.735,0.90) (36.680,-0.10)};
\nextgroupplot[title={miniImageNet},yticklabels={,,}]
\draw[densely dashed,black!55,line width=0.45pt] (axis cs:0,-0.42)--(axis cs:0,2.42);
\draw[blue!70,line width=1.05pt] (axis cs:0,2.10)--(axis cs:9.216,2.10);
\draw[orange!88!black,line width=1.05pt,-{Latex[length=2.2pt,width=1.7pt]},shorten <=2.2pt,shorten >=2.2pt] (axis cs:9.216,2.10)--(axis cs:14.291,1.90);
\draw[blue!85!black,line width=0.55pt] (axis cs:6.691,2.10)--(axis cs:11.985,2.10);
\draw[blue!85!black,line width=0.55pt] (axis cs:6.691,2.035)--(axis cs:6.691,2.165);
\draw[blue!85!black,line width=0.55pt] (axis cs:11.985,2.035)--(axis cs:11.985,2.165);
\draw[orange!88!black,line width=0.55pt] (axis cs:10.442,1.90)--(axis cs:18.607,1.90);
\draw[orange!88!black,line width=0.55pt] (axis cs:10.442,1.835)--(axis cs:10.442,1.965);
\draw[orange!88!black,line width=0.55pt] (axis cs:18.607,1.835)--(axis cs:18.607,1.965);
\draw[blue!70,line width=1.05pt] (axis cs:0,1.10)--(axis cs:10.262,1.10);
\draw[orange!88!black,line width=1.05pt,-{Latex[length=2.2pt,width=1.7pt]},shorten <=2.2pt,shorten >=2.2pt] (axis cs:10.262,1.10)--(axis cs:16.212,0.90);
\draw[blue!85!black,line width=0.55pt] (axis cs:7.398,1.10)--(axis cs:13.373,1.10);
\draw[blue!85!black,line width=0.55pt] (axis cs:7.398,1.035)--(axis cs:7.398,1.165);
\draw[blue!85!black,line width=0.55pt] (axis cs:13.373,1.035)--(axis cs:13.373,1.165);
\draw[orange!88!black,line width=0.55pt] (axis cs:12.211,0.90)--(axis cs:20.528,0.90);
\draw[orange!88!black,line width=0.55pt] (axis cs:12.211,0.835)--(axis cs:12.211,0.965);
\draw[orange!88!black,line width=0.55pt] (axis cs:20.528,0.835)--(axis cs:20.528,0.965);
\draw[blue!70,line width=1.05pt] (axis cs:0,0.10)--(axis cs:12.492,0.10);
\draw[orange!88!black,line width=1.05pt,-{Latex[length=2.2pt,width=1.7pt]},shorten <=2.2pt,shorten >=2.2pt] (axis cs:12.492,0.10)--(axis cs:20.067,-0.10);
\draw[blue!85!black,line width=0.55pt] (axis cs:9.505,0.10)--(axis cs:15.703,0.10);
\draw[blue!85!black,line width=0.55pt] (axis cs:9.505,0.035)--(axis cs:9.505,0.165);
\draw[blue!85!black,line width=0.55pt] (axis cs:15.703,0.035)--(axis cs:15.703,0.165);
\draw[orange!88!black,line width=0.55pt] (axis cs:15.802,-0.10)--(axis cs:24.615,-0.10);
\draw[orange!88!black,line width=0.55pt] (axis cs:15.802,-0.165)--(axis cs:15.802,-0.035);
\draw[orange!88!black,line width=0.55pt] (axis cs:24.615,-0.165)--(axis cs:24.615,-0.035);
\addplot+[forget plot,only marks,mark=o,mark size=1.65pt,draw=black!65,mark options={solid,fill=white}] coordinates {(0,2.10) (0,1.10) (0,0.10)};
\addplot+[only marks,mark=*,mark size=2.0pt,blue!85!black] coordinates {(9.216,2.10) (10.262,1.10) (12.492,0.10)};
\addplot+[only marks,mark=square*,mark size=2.0pt,orange!88!black] coordinates {(14.291,1.90) (16.212,0.90) (20.067,-0.10)};
\end{groupplot}
\node[anchor=north,font=\small] at ($(group c2r1.south)+(0,-0.48cm)$) {Novel-accuracy change from five-shot cosine (points)};
\node[anchor=north] at ($(group c2r1.south)+(0,-0.92cm)$) {\pgfplotslegendfromname{decomposition-legend}};
\end{tikzpicture}
	\caption{Changes in class-balanced novel accuracy at the final session after
		replacing novel means and then changing the classifier, across three datasets
		and three CLIP encoders. Each row starts at zero
		with five-shot cosine NCM. The blue point replaces only novel means with means
		from the complete training split. The orange point then switches to a linear
		probe fitted to the same training images. A leftward arrow means the probe
		performs worse than complete-mean cosine. Whiskers show 95\% two-way bootstrap
		intervals over support draws and novel classes. This breakdown follows the specified comparison order. Its components should not be interpreted as causal contributions.}
	\label{fig:decomposition}
\end{figure*}
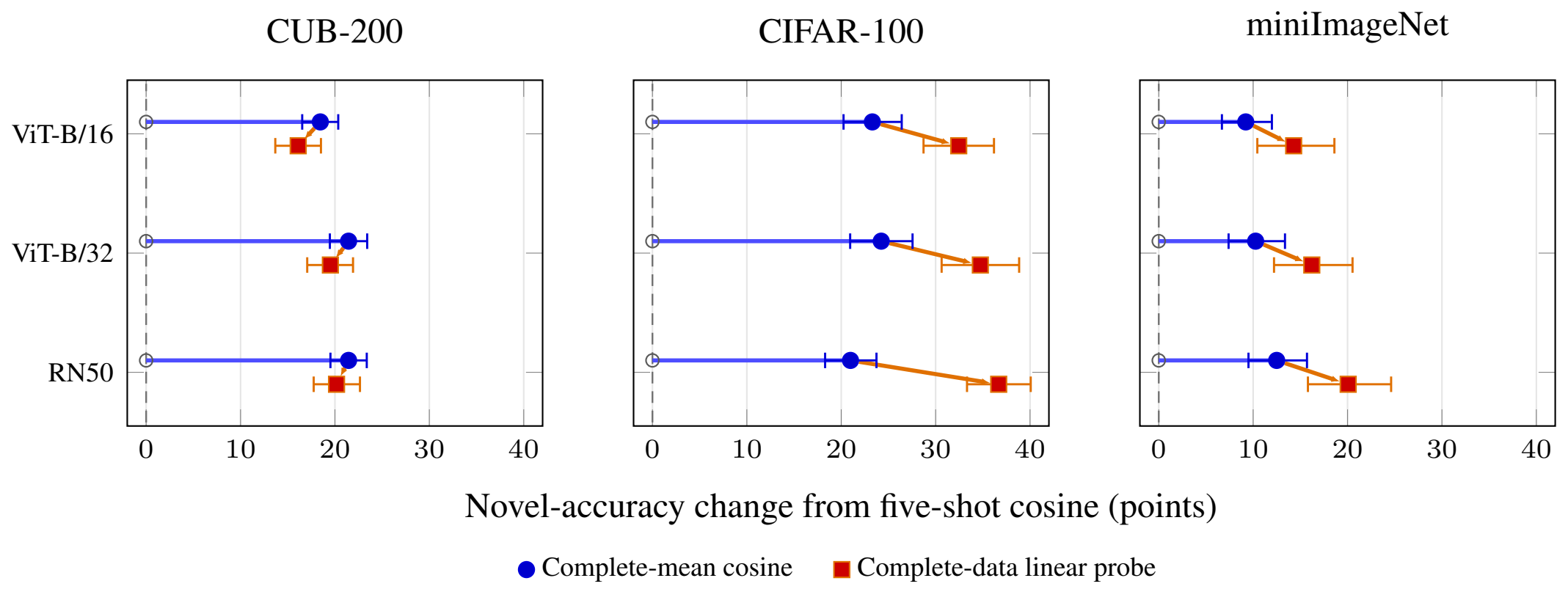


\begin{table*}[!t]
\centering
\caption{Session-wise accuracy of support-only methods on CUB-200 using frozen
	CLIP ViT-B/16 features. Avg. is the mean accuracy across all sessions, and PD
	is the performance degradation. \textbf{Bold} and \underline{underlined } values indicate the best and second-best results,
	respectively. $^{\dagger}$HyCal denotes the image-only algebraic hard-threshold
	reduction evaluated here.}
\label{tab:inductive}
\scriptsize
\setlength{\tabcolsep}{2.7pt}
\resizebox{\textwidth}{!}{%
\begin{tabular}{l*{13}{c}}
\toprule
Method & \multicolumn{11}{c}{Accuracy in each session (\%) $\uparrow$} & Avg. $\uparrow$ & PD $\downarrow$ \\
\cmidrule(lr){2-12}
 & 0 & 1 & 2 & 3 & 4 & 5 & 6 & 7 & 8 & 9 & 10 & & \\
\midrule
Cosine-NCM & 81.63 & 79.14 & 77.49 & 74.25 & 73.26 & 71.69 & 70.81 & 70.04 & 68.53 & 68.72 & 68.48 & 73.09 & 13.16 \\
TEEN (released) & 81.63 & 78.54 & 76.64 & 72.90 & 71.85 & 70.33 & 69.50 & 68.68 & 67.08 & 67.26 & 67.09 & 71.95 & 14.54 \\
ULA (released) & 81.63 & 79.18 & 77.57 & 74.30 & 73.20 & 71.63 & 70.93 & 70.16 & 68.61 & 68.83 & 68.62 & 73.15 & 13.01 \\
FeCAM & 77.06 & 71.92 & 67.35 & 63.21 & 60.64 & 58.29 & 55.68 & 53.69 & 51.68 & 50.81 & 49.71 & 60.00 & 27.35 \\
HyCal image-only reduction$^{\dagger}$ & 79.78 & 74.06 & 69.27 & 64.92 & 62.10 & 59.66 & 56.85 & 54.87 & 52.70 & 51.95 & 50.80 & 61.54 & 28.98 \\
RanDumb (RBF-NCM) & 81.56 & 78.96 & 77.21 & 74.02 & 72.93 & 71.35 & 70.47 & 69.70 & 68.22 & 68.35 & 68.08 & 72.80 & 13.49 \\
RanPAC (10k projection) & 83.62 & 79.48 & 76.09 & 72.65 & 70.96 & 68.77 & 66.86 & 65.71 & 63.86 & 63.75 & 63.09 & 70.44 & 20.54 \\
Shared-covariance LDA & \textbf{84.88} & \textbf{81.85} & \underline{80.03} & \underline{77.12} & \underline{76.23} & \underline{74.70} & \underline{73.80} & \underline{73.34} & \underline{72.14} & \underline{72.32} & \underline{72.20} & \underline{76.24} & \underline{12.68} \\
\ours{} & \underline{84.25} & \underline{81.71} & \textbf{80.23} & \textbf{77.36} & \textbf{76.50} & \textbf{75.08} & \textbf{74.38} & \textbf{73.87} & \textbf{72.62} & \textbf{72.83} & \textbf{72.72} & \textbf{76.50} & \textbf{11.53} \\
\bottomrule
\end{tabular}%
}
\end{table*}

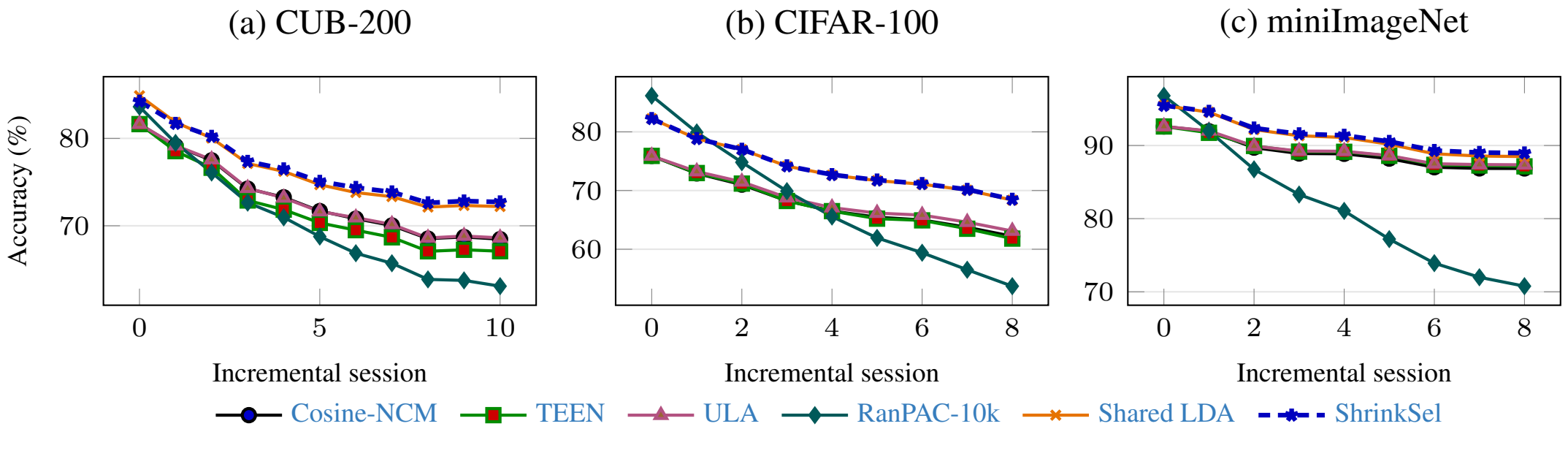
\begin{figure*}[!t]
	\centering

\begin{tikzpicture}
\begin{groupplot}[group style={group size=3 by 1,horizontal sep=0.72cm},width=0.315\textwidth,height=3.65cm,
xlabel={Incremental session},ymajorgrids=true,grid style={black!10},
tick label style={font=\scriptsize},label style={font=\scriptsize},title style={font=\small},
legend style={font=\scriptsize,draw=none,legend columns=6,/tikz/every even column/.append style={column sep=0.12cm}}]
\nextgroupplot[title={(a) CUB-200},ylabel={Accuracy (\%)},legend to name=session-curve-legend]
\addplot+[color=black,mark=*,mark size=1.7pt,thick] coordinates {(0,81.634) (1,79.139) (2,77.493) (3,74.249) (4,73.265) (5,71.686) (6,70.815) (7,70.038) (8,68.532) (9,68.720) (10,68.475)};
\addlegendentry{Cosine-NCM}
\addplot+[color=green!55!black,mark=square*,mark size=1.7pt,thick] coordinates {(0,81.634) (1,78.540) (2,76.640) (3,72.896) (4,71.852) (5,70.329) (6,69.498) (7,68.679) (8,67.080) (9,67.255) (10,67.093)};
\addlegendentry{TEEN}
\addplot+[color=magenta!70!black,mark=triangle*,mark size=1.7pt,thick] coordinates {(0,81.634) (1,79.178) (2,77.573) (3,74.300) (4,73.195) (5,71.630) (6,70.930) (7,70.159) (8,68.606) (9,68.825) (10,68.619)};
\addlegendentry{ULA}
\addplot+[color=teal!70!black,mark=diamond*,mark size=1.7pt,thick] coordinates {(0,83.624) (1,79.483) (2,76.095) (3,72.648) (4,70.962) (5,68.766) (6,66.860) (7,65.713) (8,63.865) (9,63.751) (10,63.089)};
\addlegendentry{RanPAC-10k}
\addplot+[color=orange!90!black,mark=x,mark size=1.7pt,thick] coordinates {(0,84.881) (1,81.845) (2,80.035) (3,77.123) (4,76.225) (5,74.697) (6,73.803) (7,73.336) (8,72.140) (9,72.322) (10,72.202)};
\addlegendentry{Shared LDA}
\addplot+[color=blue!75!black,mark=star,mark size=1.7pt,very thick] coordinates {(0,84.253) (1,81.707) (2,80.226) (3,77.361) (4,76.504) (5,75.077) (6,74.377) (7,73.868) (8,72.620) (9,72.832) (10,72.724)};
\addlegendentry{ShrinkSel}
\nextgroupplot[title={(b) CIFAR-100}]
\addplot+[color=black,mark=*,mark size=1.7pt,thick] coordinates {(0,75.900) (1,72.908) (2,71.006) (3,68.241) (4,66.506) (5,65.458) (6,64.987) (7,63.736) (8,62.207)};
\addplot+[color=green!55!black,mark=square*,mark size=1.7pt,thick] coordinates {(0,75.900) (1,72.979) (2,71.184) (3,68.221) (4,66.489) (5,65.209) (6,64.919) (7,63.490) (8,61.811)};
\addplot+[color=magenta!70!black,mark=triangle*,mark size=1.7pt,thick] coordinates {(0,75.900) (1,73.235) (2,71.485) (3,68.791) (4,67.089) (5,66.151) (6,65.813) (7,64.606) (8,63.103)};
\addplot+[color=teal!70!black,mark=diamond*,mark size=1.7pt,thick] coordinates {(0,86.150) (1,79.905) (2,74.814) (3,69.897) (4,65.569) (5,61.932) (6,59.404) (7,56.503) (8,53.705)};
\addplot+[color=orange!90!black,mark=x,mark size=1.7pt,thick] coordinates {(0,82.250) (1,78.820) (2,76.992) (3,74.153) (4,72.638) (5,71.694) (6,71.071) (7,70.070) (8,68.373)};
\addplot+[color=blue!75!black,mark=star,mark size=1.7pt,very thick] coordinates {(0,82.250) (1,78.835) (2,77.018) (3,74.191) (4,72.701) (5,71.779) (6,71.174) (7,70.184) (8,68.496)};
\nextgroupplot[title={(c) miniImageNet}]
\addplot+[color=black,mark=*,mark size=1.7pt,thick] coordinates {(0,92.600) (1,91.945) (2,89.708) (3,88.903) (4,88.863) (5,88.206) (6,87.028) (7,86.868) (8,86.861)};
\addplot+[color=green!55!black,mark=square*,mark size=1.7pt,thick] coordinates {(0,92.600) (1,91.733) (2,89.939) (3,89.161) (4,89.137) (5,88.589) (6,87.353) (7,87.267) (8,87.147)};
\addplot+[color=magenta!70!black,mark=triangle*,mark size=1.7pt,thick] coordinates {(0,92.600) (1,92.004) (2,89.915) (3,89.237) (4,89.213) (5,88.613) (6,87.503) (7,87.390) (8,87.360)};
\addplot+[color=teal!70!black,mark=diamond*,mark size=1.7pt,thick] coordinates {(0,96.800) (1,91.976) (2,86.739) (3,83.293) (4,81.059) (5,77.201) (6,73.913) (7,71.986) (8,70.767)};
\addplot+[color=orange!90!black,mark=x,mark size=1.7pt,thick] coordinates {(0,95.467) (1,94.554) (2,92.168) (3,91.329) (4,91.094) (5,90.168) (6,88.867) (7,88.556) (8,88.489)};
\addplot+[color=blue!75!black,mark=star,mark size=1.7pt,very thick] coordinates {(0,95.467) (1,94.661) (2,92.378) (3,91.587) (4,91.386) (5,90.545) (6,89.307) (7,89.037) (8,88.970)};
\end{groupplot}
\node[anchor=north] at ($(group c2r1.south)+(0,-0.62cm)$) {\ref{session-curve-legend}};
\end{tikzpicture}
	\caption{Session-wise accuracy of support-only methods on (a) CUB-200,
		(b) CIFAR-100, and (c) miniImageNet using frozen CLIP ViT-B/16 features.}
	\label{fig:session-curves}
\end{figure*}

\begin{figure*}[!t]
	\centering
\begin{tikzpicture}
\begin{groupplot}[
group style={group size=3 by 1,horizontal sep=0.72cm},
width=0.285\textwidth,height=4.55cm,
xmin=-0.5,xmax=6.5,ymin=-0.5,ymax=2.5,
xtick={0,1,2,3,4,5,6},xticklabels={$0$,$.1$,$.25$,$.5$,$.75$,$.9$,$1$},
ytick={0,1,2},yticklabels={unit,raw,corrected},
xlabel={Covariance shrinkage $\lambda$},
tick label style={font=\scriptsize},label style={font=\scriptsize},title style={font=\small},
colormap={seldiv}{color(0cm)=(orange!60);color(5cm)=(white);color(6.25cm)=(blue!60)},
point meta min=-2,point meta max=1.25,axis on top,
]
\nextgroupplot[title={CUB-200}]
\addplot[matrix plot*,mesh/cols=7,point meta=explicit] table[meta=value] {
x y value
0 0 0.126313955
1 0 0.391238143
2 0 0.604858046
3 0 0.870337960
4 0 1.023428823
5 0 0.378304009
6 0 -1.938472912
0 1 0.000000000
1 1 0.297386304
2 1 0.608767395
3 1 0.988323676
4 1 1.072096727
5 1 0.493056466
6 1 -1.991162634
0 2 0.153598463
1 2 0.453153195
2 2 0.689619149
3 2 1.020509682
4 2 1.136149650
5 2 0.569410245
6 2 -1.915887327
};
\node[anchor=north east,font=\tiny,text=black!55] at (axis cs:0.43,0.43) {$\times$};
\node[anchor=north east,font=\tiny,text=black!55] at (axis cs:1.43,0.43) {$\times$};
\node[anchor=north east,font=\tiny,text=black!55] at (axis cs:2.43,0.43) {$\times$};
\draw[black!65,densely dashed,line width=0.55pt] (axis cs:2.54,0.54) rectangle (axis cs:3.46,1.46);
\draw[black!65,densely dashed,line width=0.55pt] (axis cs:2.54,1.54) rectangle (axis cs:3.46,2.46);
\draw[black!65,densely dashed,line width=0.55pt] (axis cs:3.54,-0.46) rectangle (axis cs:4.46,0.46);
\draw[black!65,densely dashed,line width=0.55pt] (axis cs:3.54,0.54) rectangle (axis cs:4.46,1.46);
\draw[black,line width=1.0pt] (axis cs:3.53,1.53) rectangle (axis cs:4.47,2.47);
\node[anchor=north east,font=\tiny,text=black!55] at (axis cs:5.43,0.43) {$\times$};
\node[anchor=north east,font=\tiny,text=black!55] at (axis cs:5.43,1.43) {$\times$};
\node[anchor=north east,font=\tiny,text=black!55] at (axis cs:5.43,2.43) {$\times$};
\node[anchor=north east,font=\tiny,text=black!55] at (axis cs:6.43,0.43) {$\times$};
\node[anchor=north east,font=\tiny,text=black!55] at (axis cs:6.43,1.43) {$\times$};
\node[anchor=north east,font=\tiny,text=black!55] at (axis cs:6.43,2.43) {$\times$};
\nextgroupplot[title={CIFAR-100},yticklabels={,,}]
\addplot[matrix plot*,mesh/cols=7,point meta=explicit] table[meta=value] {
x y value
0 0 0.113398792
1 0 0.128438042
2 0 0.030413517
3 0 -0.417847750
4 0 -0.949112975
5 0 -1.746436746
6 0 -4.773248734
0 1 0.000000000
1 1 -0.095106796
2 1 -0.006312216
3 1 -0.136504923
4 1 -0.725143112
5 1 -1.939098595
6 1 -4.845990135
0 2 0.101297336
1 2 0.010987406
2 2 0.085400866
3 2 -0.037957948
4 2 -0.625263048
5 2 -1.824462868
6 2 -4.777932436
};
\node[anchor=north east,font=\tiny,text=black!55] at (axis cs:0.43,0.43) {$\times$};
\draw[black!65,densely dashed,line width=0.55pt] (axis cs:-0.46,0.54) rectangle (axis cs:0.46,1.46);
\draw[black,line width=1.0pt] (axis cs:-0.47,1.53) rectangle (axis cs:0.47,2.47);
\node[anchor=north east,font=\tiny,text=black!55] at (axis cs:1.43,0.43) {$\times$};
\node[anchor=north east,font=\tiny,text=black!55] at (axis cs:1.43,1.43) {$\times$};
\node[anchor=north east,font=\tiny,text=black!55] at (axis cs:1.43,2.43) {$\times$};
\node[anchor=north east,font=\tiny,text=black!55] at (axis cs:2.43,0.43) {$\times$};
\node[anchor=north east,font=\tiny,text=black!55] at (axis cs:2.43,1.43) {$\times$};
\node[anchor=north east,font=\tiny,text=black!55] at (axis cs:2.43,2.43) {$\times$};
\node[anchor=north east,font=\tiny,text=black!55] at (axis cs:3.43,0.43) {$\times$};
\node[anchor=north east,font=\tiny,text=black!55] at (axis cs:3.43,1.43) {$\times$};
\node[anchor=north east,font=\tiny,text=black!55] at (axis cs:3.43,2.43) {$\times$};
\node[anchor=north east,font=\tiny,text=black!55] at (axis cs:4.43,0.43) {$\times$};
\node[anchor=north east,font=\tiny,text=black!55] at (axis cs:4.43,1.43) {$\times$};
\node[anchor=north east,font=\tiny,text=black!55] at (axis cs:4.43,2.43) {$\times$};
\node[anchor=north east,font=\tiny,text=black!55] at (axis cs:5.43,0.43) {$\times$};
\node[anchor=north east,font=\tiny,text=black!55] at (axis cs:5.43,1.43) {$\times$};
\node[anchor=north east,font=\tiny,text=black!55] at (axis cs:5.43,2.43) {$\times$};
\node[anchor=north east,font=\tiny,text=black!55] at (axis cs:6.43,0.43) {$\times$};
\node[anchor=north east,font=\tiny,text=black!55] at (axis cs:6.43,1.43) {$\times$};
\node[anchor=north east,font=\tiny,text=black!55] at (axis cs:6.43,2.43) {$\times$};
\nextgroupplot[title={miniImageNet},yticklabels={,,},colorbar,colorbar style={width=0.12cm,ylabel={Mean gain over shared LDA (points)},ylabel style={font=\scriptsize},tick label style={font=\tiny}}]
\addplot[matrix plot*,mesh/cols=7,point meta=explicit] table[meta=value] {
x y value
0 0 0.024435296
1 0 -0.038275594
2 0 0.047007256
3 0 0.051870970
4 0 -0.380372373
5 0 -0.878786047
6 0 -2.137873059
0 1 0.000000000
1 1 -0.029107349
2 1 -0.035544868
3 1 -0.099590702
4 1 -0.617997716
5 1 -1.218759384
6 1 -2.414494541
0 2 0.278283635
1 2 0.278248378
2 2 0.257429732
3 2 0.199504401
4 2 -0.326346623
5 2 -1.001670716
6 2 -2.183935099
};
\node[anchor=north east,font=\tiny,text=black!55] at (axis cs:0.43,0.43) {$\times$};
\draw[black,line width=1.0pt] (axis cs:-0.47,1.53) rectangle (axis cs:0.47,2.47);
\node[anchor=north east,font=\tiny,text=black!55] at (axis cs:1.43,0.43) {$\times$};
\node[anchor=north east,font=\tiny,text=black!55] at (axis cs:1.43,1.43) {$\times$};
\draw[black!65,densely dashed,line width=0.55pt] (axis cs:0.54,1.54) rectangle (axis cs:1.46,2.46);
\node[anchor=north east,font=\tiny,text=black!55] at (axis cs:2.43,0.43) {$\times$};
\node[anchor=north east,font=\tiny,text=black!55] at (axis cs:2.43,1.43) {$\times$};
\draw[black!65,densely dashed,line width=0.55pt] (axis cs:1.54,1.54) rectangle (axis cs:2.46,2.46);
\node[anchor=north east,font=\tiny,text=black!55] at (axis cs:3.43,0.43) {$\times$};
\node[anchor=north east,font=\tiny,text=black!55] at (axis cs:3.43,1.43) {$\times$};
\node[anchor=north east,font=\tiny,text=black!55] at (axis cs:3.43,2.43) {$\times$};
\node[anchor=north east,font=\tiny,text=black!55] at (axis cs:4.43,0.43) {$\times$};
\node[anchor=north east,font=\tiny,text=black!55] at (axis cs:4.43,1.43) {$\times$};
\node[anchor=north east,font=\tiny,text=black!55] at (axis cs:4.43,2.43) {$\times$};
\node[anchor=north east,font=\tiny,text=black!55] at (axis cs:5.43,0.43) {$\times$};
\node[anchor=north east,font=\tiny,text=black!55] at (axis cs:5.43,1.43) {$\times$};
\node[anchor=north east,font=\tiny,text=black!55] at (axis cs:5.43,2.43) {$\times$};
\node[anchor=north east,font=\tiny,text=black!55] at (axis cs:6.43,0.43) {$\times$};
\node[anchor=north east,font=\tiny,text=black!55] at (axis cs:6.43,1.43) {$\times$};
\node[anchor=north east,font=\tiny,text=black!55] at (axis cs:6.43,2.43) {$\times$};
\end{groupplot}
\end{tikzpicture}
	\caption{Base-only selection of prototype radius and covariance shrinkage
		for \ours{} using CLIP ViT-B/16 features. Colors show the mean schedule-weighted
		accuracy gain over shared LDA across class-disjoint base folds. A $\times$
		marks a candidate that fails the non-regression constraint. Solid boxes
		indicate the selected candidates; dashed boxes indicate other feasible
		candidates within $0.15$ point of the selected objective. Gains below $-2$
		percentage points are clipped for display.}
	\label{fig:selector-landscape}
\end{figure*}
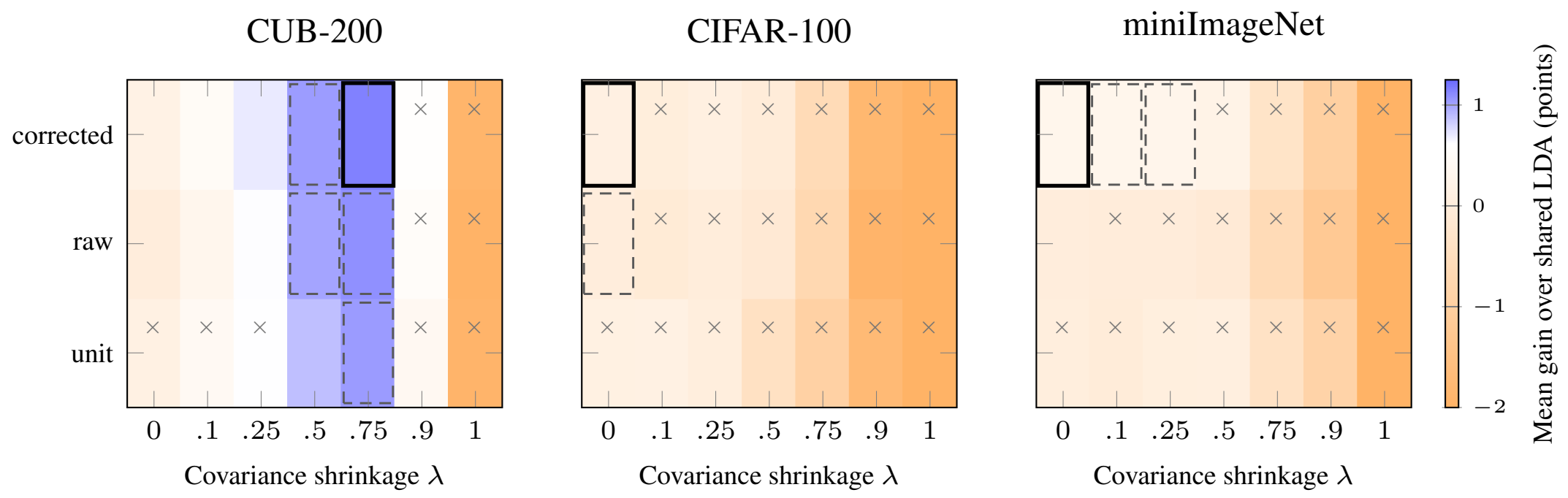

\begin{table*}[!t]
\centering
\caption{Query-assisted methods compared with cosine NCM and ShrinkSel as support-only references, using frozen CLIP ViT-B/16 features and twenty matched support draws. The support-only row in each block gives the reference accuracy. Avg. is session-average accuracy, and $\Delta$ Parent is the paired gain over that reference, with a 95\% bootstrap interval. Settings chosen on base classes are kept fixed when changing heads. Bold and underline mark the best and second-best displayed accuracy and gain within each block; displayed ties share formatting. $^{\dagger}$These variants start from ShrinkSel prototypes in its whitened geometry and retain each method's updates, including full LaplacianShot's CL2N preprocessing. MAP-OT$_u$ assumes a uniform query marginal.}
\label{tab:transductive-benchmark}
\small
\setlength{\tabcolsep}{3.0pt}
\begin{tabular*}{\textwidth}{@{\extracolsep{\fill}}l*{3}{cc}@{}}
\toprule
Method & \multicolumn{2}{c}{CUB-200} & \multicolumn{2}{c}{CIFAR-100} & \multicolumn{2}{c}{miniImageNet} \\
\cmidrule(lr){2-3}\cmidrule(lr){4-5}\cmidrule(lr){6-7}
 & Avg. $\uparrow$ & $\Delta$ Parent [95\%] $\uparrow$ & Avg. $\uparrow$ & $\Delta$ Parent [95\%] $\uparrow$ & Avg. $\uparrow$ & $\Delta$ Parent [95\%] $\uparrow$ \\
\midrule
\multicolumn{7}{l}{\textit{Reference head: Cosine-NCM}} \\
Support-only & 73.09 & --- & 67.88 & --- & 89.00 & --- \\
\midrule
QPR & \underline{73.76} & \underline{+0.67 [+0.60, +0.73]} & \underline{68.25} & \underline{+0.37 [+0.32, +0.42]} & \underline{89.54} & \underline{+0.54 [+0.51, +0.58]} \\
QPR (self-excl.) & \textbf{73.83} & \textbf{+0.73 [+0.68, +0.79]} & 68.24 & +0.36 [+0.31, +0.41] & \underline{89.54} & \underline{+0.54 [+0.51, +0.58]} \\
TIM-GD & 72.99 & -0.11 [-0.22, 0.00] & \textbf{69.40} & \textbf{+1.51 [+1.43, +1.60]} & \textbf{90.33} & \textbf{+1.33 [+1.28, +1.38]} \\
LapShot-full & 69.71 & -3.38 [-3.49, -3.27] & 63.18 & -4.71 [-4.84, -4.57] & 86.02 & -2.98 [-3.03, -2.92] \\
LapShot-core & 73.03 & -0.06 [-0.07, -0.06] & 67.80 & -0.08 [-0.09, -0.07] & 89.03 & +0.03 [+0.02, +0.04] \\
ProtoLP & 66.90 & -6.20 [-6.43, -5.97] & 62.75 & -5.14 [-5.24, -5.03] & 86.71 & -2.28 [-2.38, -2.18] \\
\addlinespace[2pt]
MAP-OT$_u$ & 58.13 & -14.96 [-15.11, -14.80] & 59.35 & -8.53 [-8.64, -8.42] & 78.35 & -10.65 [-10.84, -10.46] \\
\midrule
\multicolumn{7}{l}{\textit{Reference head: ShrinkSel}} \\
Support-only & 76.50 & --- & 74.07 & --- & 91.48 & --- \\
\midrule
QPR & \underline{78.13} & \underline{+1.63 [+1.53, +1.71]} & \underline{76.10} & \underline{+2.03 [+1.93, +2.15]} & \textbf{92.69} & \textbf{+1.21 [+1.15, +1.28]} \\
QPR (self-excl.) & \textbf{78.35} & \textbf{+1.84 [+1.73, +1.94]} & \underline{76.10} & \underline{+2.03 [+1.91, +2.14]} & \underline{92.67} & \underline{+1.19 [+1.12, +1.26]} \\
TIM-GD\textsuperscript{$\dagger$} & 74.78 & -1.72 [-1.87, -1.58] & \textbf{76.37} & \textbf{+2.30 [+2.18, +2.41]} & 92.59 & +1.11 [+1.03, +1.19] \\
LapShot-full\textsuperscript{$\dagger$} & 76.66 & +0.16 [+0.09, +0.22] & 72.87 & -1.20 [-1.31, -1.08] & 91.37 & -0.11 [-0.19, -0.03] \\
LapShot-core\textsuperscript{$\dagger$} & 76.54 & +0.03 [+0.03, +0.04] & 74.05 & -0.02 [-0.03, -0.02] & 91.50 & +0.01 [0.00, +0.02] \\
ProtoLP\textsuperscript{$\dagger$} & 71.56 & -4.94 [-5.15, -4.74] & 71.81 & -2.26 [-2.41, -2.11] & 90.61 & -0.87 [-0.99, -0.76] \\
\addlinespace[2pt]
MAP-OT$_u$\textsuperscript{$\dagger$} & 59.03 & -17.48 [-17.72, -17.21] & 70.84 & -3.23 [-3.40, -3.06] & 89.67 & -1.81 [-1.95, -1.67] \\
\bottomrule
\end{tabular*}
\end{table*}

\begin{figure*}[!t]
\centering
\begin{minipage}[t]{\columnwidth}
\vspace{0pt}
	\centering
\begin{tikzpicture}
\begin{groupplot}[
group style={group size=1 by 2,vertical sep=1.75cm},
width=0.98\columnwidth,height=3.7cm,
xmin=-0.48,xmax=4.48,ymin=-1.650,ymax=9.150,
xtick={0,1,2,3,4},ytick distance=2,
ymajorgrids=true,grid style={black!10},axis line style={black!45},
tick label style={font=\small},label style={font=\small},title style={font=\small},
legend style={font=\small,draw=none,legend columns=3},
ylabel={QPR gain (points)},
]
\nextgroupplot[title={(a) Labeled supports},xlabel={$K$ labeled supports},xticklabels={1,2,5,10,20},legend to name=qpr-boundary-legend]
\addplot+[only marks,mark=*,mark size=2.15pt,color=blue!75!black,error bars/.cd,y dir=both,y explicit] table[x=x,y=mean,y error plus=plus,y error minus=minus] {
x mean plus minus
-0.140 5.474629 0.290840 0.276148
0.860 6.112358 0.217488 0.194188
1.860 2.525026 0.139799 0.138937
2.860 0.874180 0.076803 0.075078
3.860 0.182948 0.053504 0.051800
};
\addlegendentry{CUB-200}
\addplot+[only marks,mark=square*,mark size=2.15pt,color=orange!90!black,error bars/.cd,y dir=both,y explicit] table[x=x,y=mean,y error plus=plus,y error minus=minus] {
x mean plus minus
0.000 3.588001 0.319500 0.307513
1.000 5.881000 0.218999 0.219000
2.000 3.549001 0.177500 0.169500
3.000 1.319000 0.105500 0.102500
4.000 0.136500 0.074500 0.074001
};
\addlegendentry{CIFAR-100}
\addplot+[only marks,mark=triangle*,mark size=2.15pt,color=teal!70!black,error bars/.cd,y dir=both,y explicit] table[x=x,y=mean,y error plus=plus,y error minus=minus] {
x mean plus minus
0.140 8.172000 0.512013 0.516499
1.140 8.184500 0.343999 0.324000
2.140 2.175001 0.136000 0.129000
3.140 0.638001 0.077499 0.075999
4.140 0.080501 0.054000 0.048499
};
\addlegendentry{miniImageNet}
\draw[densely dashed,black!65] (axis cs:-0.48,0) -- (axis cs:4.48,0);
\nextgroupplot[title={(b) Balanced peer-query cohort},xlabel={Queries per seen class},xticklabels={1,2,5,10,20}]
\addplot+[only marks,mark=*,mark size=2.15pt,color=blue!75!black,error bars/.cd,y dir=both,y explicit] table[x=x,y=mean,y error plus=plus,y error minus=minus] {
x mean plus minus
-0.140 -0.025000 0.025000 0.050000
0.860 -0.125000 0.124999 0.112500
1.860 -0.905000 0.249999 0.259999
2.860 1.565000 0.140000 0.137500
3.860 2.318331 0.144614 0.144581
};
\addplot+[only marks,mark=square*,mark size=2.15pt,color=orange!90!black,error bars/.cd,y dir=both,y explicit] table[x=x,y=mean,y error plus=plus,y error minus=minus] {
x mean plus minus
0.000 0.000000 0.000000 0.000000
1.000 0.000000 0.000000 0.000000
2.000 0.040000 0.110000 0.090000
3.000 1.350000 0.135001 0.135000
4.000 2.474999 0.147500 0.140000
};
\addplot+[only marks,mark=triangle*,mark size=2.15pt,color=teal!70!black,error bars/.cd,y dir=both,y explicit] table[x=x,y=mean,y error plus=plus,y error minus=minus] {
x mean plus minus
0.140 0.000000 0.000000 0.000000
1.140 0.000000 0.000000 0.000000
2.140 -0.070000 0.060000 0.070000
3.140 0.715000 0.100001 0.104999
4.140 1.460000 0.157500 0.162500
};
\draw[densely dashed,black!65] (axis cs:-0.48,0) -- (axis cs:4.48,0);
\end{groupplot}
\node[anchor=north] at ($(group c1r2.south west)!0.5!(group c1r2.south east)+(0,-1.0cm)$) {\ref{qpr-boundary-legend}};
\end{tikzpicture}
	\caption{How QPR's gain changes with labeled support and query-cohort size on
		CLIP ViT-B/16. The top panel uses the complete all-seen query cohort and varies
		the number of labeled supports. The bottom fixes $K=5$ and varies the number of
		queries per class in a balanced cohort. Both panels show the final-session gain
		in overall (Full) accuracy over \ours{}. Error bars give paired 95\% bootstrap
		intervals computed from twenty support draws. The same base-selected QPR
		settings are used throughout.}
	\label{fig:qpr-operating-boundary}
\end{minipage}\hfill
\begin{minipage}[t]{\columnwidth}
\vspace{0pt}
	\centering
\begin{tikzpicture}
\begin{groupplot}[
group style={group size=1 by 3,vertical sep=0.75cm,xlabels at=edge bottom,xticklabels at=edge bottom},
scale only axis,width=0.66\columnwidth,height=1.95cm,
xmin=-0.5,xmax=4.5,ymin=-0.5,ymax=4.5,
xtick={0,1,2,3,4},xticklabels={$1$,$2$,$5$,$10$,$20$},
ytick={0,1,2,3,4},yticklabels={$1$,$2$,$5$,$10$,$20$},
xlabel={Queries per seen class},ylabel={$K$ labeled supports},
tick label style={font=\small},label style={font=\small},title style={font=\small},
colormap={qprgrid}{color(0cm)=(orange!65);color(1cm)=(white);color(2cm)=(blue!65)},
point meta min=-5.500000,point meta max=5.500000,axis on top,
]
\nextgroupplot[title={CUB-200},ylabel={}]
\addplot[matrix plot*,mesh/cols=5,point meta=explicit] table[meta=value] {
x y value
0 0 0.000000000
1 0 -0.112500042
2 0 0.154999346
3 0 1.719999611
4 0 4.205431491
0 1 0.025000274
1 1 -0.450000465
2 1 -0.890000165
3 1 2.665000856
4 1 5.237616599
0 2 -0.024999976
1 2 -0.125000477
2 2 -0.904999971
3 2 1.564999819
4 2 2.318330705
0 3 0.000000000
1 3 -0.062499940
2 3 -0.460000634
3 3 0.725000203
4 3 0.841085911
0 4 0.000000000
1 4 -0.024999976
2 4 -0.304999948
3 4 0.234999955
4 4 0.188584924
};
\node[font=\small,text=black] at (axis cs:0,0) {0.0};
\node[font=\small,text=black] at (axis cs:1,0) {-0.1};
\node[font=\small,text=black] at (axis cs:2,0) {+0.2};
\node[font=\small,text=black] at (axis cs:3,0) {+1.7};
\node[font=\small,text=white] at (axis cs:4,0) {+4.2};
\node[font=\small,text=black] at (axis cs:0,1) {0.0};
\node[font=\small,text=black] at (axis cs:1,1) {-0.5};
\node[font=\small,text=black] at (axis cs:2,1) {-0.9};
\node[font=\small,text=black] at (axis cs:3,1) {+2.7};
\node[font=\small,text=white] at (axis cs:4,1) {+5.2};
\node[font=\small,text=black] at (axis cs:0,2) {0.0};
\node[font=\small,text=black] at (axis cs:1,2) {-0.1};
\node[font=\small,text=black] at (axis cs:2,2) {-0.9};
\node[font=\small,text=black] at (axis cs:3,2) {+1.6};
\node[font=\small,text=black] at (axis cs:4,2) {+2.3};
\node[font=\small,text=black] at (axis cs:0,3) {0.0};
\node[font=\small,text=black] at (axis cs:1,3) {-0.1};
\node[font=\small,text=black] at (axis cs:2,3) {-0.5};
\node[font=\small,text=black] at (axis cs:3,3) {+0.7};
\node[font=\small,text=black] at (axis cs:4,3) {+0.8};
\node[font=\small,text=black] at (axis cs:0,4) {0.0};
\node[font=\small,text=black] at (axis cs:1,4) {0.0};
\node[font=\small,text=black] at (axis cs:2,4) {-0.3};
\node[font=\small,text=black] at (axis cs:3,4) {+0.2};
\node[font=\small,text=black] at (axis cs:4,4) {+0.2};
\nextgroupplot[title={CIFAR-100}]
\addplot[matrix plot*,mesh/cols=5,point meta=explicit] table[meta=value] {
x y value
0 0 0.000000000
1 0 0.000000000
2 0 0.000000000
3 0 0.240000188
4 0 0.975000262
0 1 0.000000000
1 1 0.000000000
2 1 0.040000379
3 1 0.980000198
4 1 2.820000350
0 2 0.000000000
1 2 0.000000000
2 2 0.040000081
3 2 1.349999905
4 2 2.474999428
0 3 0.000000000
1 3 0.000000000
2 3 0.040000677
3 3 0.730001032
4 3 1.007500291
0 4 0.000000000
1 4 0.000000000
2 4 -0.100000501
3 4 0.259999931
4 4 0.079999268
};
\node[font=\small,text=black] at (axis cs:0,0) {0.0};
\node[font=\small,text=black] at (axis cs:1,0) {0.0};
\node[font=\small,text=black] at (axis cs:2,0) {0.0};
\node[font=\small,text=black] at (axis cs:3,0) {+0.2};
\node[font=\small,text=black] at (axis cs:4,0) {+1.0};
\node[font=\small,text=black] at (axis cs:0,1) {0.0};
\node[font=\small,text=black] at (axis cs:1,1) {0.0};
\node[font=\small,text=black] at (axis cs:2,1) {0.0};
\node[font=\small,text=black] at (axis cs:3,1) {+1.0};
\node[font=\small,text=black] at (axis cs:4,1) {+2.8};
\node[font=\small,text=black] at (axis cs:0,2) {0.0};
\node[font=\small,text=black] at (axis cs:1,2) {0.0};
\node[font=\small,text=black] at (axis cs:2,2) {0.0};
\node[font=\small,text=black] at (axis cs:3,2) {+1.3};
\node[font=\small,text=black] at (axis cs:4,2) {+2.5};
\node[font=\small,text=black] at (axis cs:0,3) {0.0};
\node[font=\small,text=black] at (axis cs:1,3) {0.0};
\node[font=\small,text=black] at (axis cs:2,3) {0.0};
\node[font=\small,text=black] at (axis cs:3,3) {+0.7};
\node[font=\small,text=black] at (axis cs:4,3) {+1.0};
\node[font=\small,text=black] at (axis cs:0,4) {0.0};
\node[font=\small,text=black] at (axis cs:1,4) {0.0};
\node[font=\small,text=black] at (axis cs:2,4) {-0.1};
\node[font=\small,text=black] at (axis cs:3,4) {+0.3};
\node[font=\small,text=black] at (axis cs:4,4) {+0.1};
\nextgroupplot[title={miniImageNet},ylabel={},colorbar,colorbar style={width=0.12cm,height=7.35cm,at={(parent axis.right of south east)},anchor=south west,ytick={-5,0,5},ylabel={Gain (points)},ylabel style={font=\small},tick label style={font=\small}}]
\addplot[matrix plot*,mesh/cols=5,point meta=explicit] table[meta=value] {
x y value
0 0 0.000000000
1 0 0.000000000
2 0 0.000000000
3 0 1.475000679
4 0 3.670000732
0 1 0.000000000
1 1 0.000000000
2 1 -0.000000596
3 1 3.060000837
4 1 5.489999950
0 2 0.000000000
1 2 0.000000000
2 2 -0.069999695
3 2 0.714999735
4 2 1.459999979
0 3 0.000000000
1 3 0.000000000
2 3 -0.040000379
3 3 0.169999897
4 3 0.422499776
0 4 0.000000000
1 4 0.000000000
2 4 -0.039999485
3 4 -0.009998977
4 4 0.092499852
};
\node[font=\small,text=black] at (axis cs:0,0) {0.0};
\node[font=\small,text=black] at (axis cs:1,0) {0.0};
\node[font=\small,text=black] at (axis cs:2,0) {0.0};
\node[font=\small,text=black] at (axis cs:3,0) {+1.5};
\node[font=\small,text=white] at (axis cs:4,0) {+3.7};
\node[font=\small,text=black] at (axis cs:0,1) {0.0};
\node[font=\small,text=black] at (axis cs:1,1) {0.0};
\node[font=\small,text=black] at (axis cs:2,1) {0.0};
\node[font=\small,text=black] at (axis cs:3,1) {+3.1};
\node[font=\small,text=white] at (axis cs:4,1) {+5.5};
\node[font=\small,text=black] at (axis cs:0,2) {0.0};
\node[font=\small,text=black] at (axis cs:1,2) {0.0};
\node[font=\small,text=black] at (axis cs:2,2) {-0.1};
\node[font=\small,text=black] at (axis cs:3,2) {+0.7};
\node[font=\small,text=black] at (axis cs:4,2) {+1.5};
\node[font=\small,text=black] at (axis cs:0,3) {0.0};
\node[font=\small,text=black] at (axis cs:1,3) {0.0};
\node[font=\small,text=black] at (axis cs:2,3) {0.0};
\node[font=\small,text=black] at (axis cs:3,3) {+0.2};
\node[font=\small,text=black] at (axis cs:4,3) {+0.4};
\node[font=\small,text=black] at (axis cs:0,4) {0.0};
\node[font=\small,text=black] at (axis cs:1,4) {0.0};
\node[font=\small,text=black] at (axis cs:2,4) {0.0};
\node[font=\small,text=black] at (axis cs:3,4) {0.0};
\node[font=\small,text=black] at (axis cs:4,4) {+0.1};
\end{groupplot}
\end{tikzpicture}
	\caption{QPR gains for different combinations of labeled support and balanced
		query-cohort size at the final session. Each cell shows QPR minus \ours{} in
		overall (Full) accuracy, measured in percentage points. Labels are used only
		to construct the balanced cohort and are hidden from both methods during
		inference. All 75 conditions use the same QPR settings selected with five
		labeled supports. This exploratory grid does not test significance separately
		for each cell.}
	\label{fig:qpr-shot-cohort-grid}
\end{minipage}
\end{figure*}

\paragraph{Metrics and uncertainty.}
The primary metric is the unweighted average of sample-weighted session
accuracies, including session 0. The diagnosis instead uses class-balanced
novel accuracy at the final session. Paired method intervals use 10,000
support resamples; diagnostic crossed support--class intervals use 5,000
draws. These intervals quantify the resampled sources of variation under
the fixed protocol. The supplement gives further metric and implementation
details.

\subsection{Locating the recoverable error}

In every row of Fig.~\ref{fig:decomposition}, replacing five-shot novel
means recovers more than half of the gap to the complete-data probe.
The cross-family and Aircraft checks show the same pattern: mean replacement
accounts for more than half of the gap in all sixteen strong encoder--dataset settings.

The subsequent probe does not always improve accuracy. On CUB, the orange
step points left because the probe underperforms complete-data cosine.
Additional classifier flexibility is therefore not necessarily beneficial
after mean replacement.

The supporting diagnostics are consistent with a finite-sample explanation.
Accuracy improves with increasing support count, and complete-data means
repair more decisions than they damage. Angular error is associated with
classwise loss on the standard matrix, but not on Aircraft despite its
large aggregate improvement. Whether a direction error changes a label
also depends on the nearest competing boundary. The matched replacement
is our primary evidence, shot and angle diagnostics help interpret it.

\subsection{Support-only recovery}

Table~\ref{tab:inductive} compares support-only heads on CUB-200.
Shared LDA is the relevant reference because it is already a member of the
\ours{} family. \ours{} is slightly below LDA in sessions 0--1 and improves
in the remaining sessions. The session-average gains over LDA are
$+0.27$, $+0.06$, and $+0.29$ percentage points on CUB-200, CIFAR-100,
and miniImageNet, respectively.

Figure~\ref{fig:session-curves} shows the trajectories on all three
benchmarks. \ours{} and LDA separate as novel classes accumulate on CUB
and miniImageNet, while CIFAR remains nearly tied. RanPAC has a strong
base-session result but falls behind as novel classes enter. Base-session
accuracy alone does not determine the session average.

The ablations indicate that covariance-aware scoring accounts for most of
the improvement over cosine. Radius correction has a smaller effect that
depends on the representation. A mean's radius contains concentration
information, but its squared value is inflated by finite-sample estimation
(Eq.~\eqref{eq:resultant-bias}). Giving a direction more weight in the metric
also increases the contribution of estimation noise along that direction, as
Equation~\eqref{eq:finite-penalty} shows. We therefore select the radius treatment
and metric together.

Figure~\ref{fig:selector-landscape} shows how performance and foldwise
feasibility determine selection. Several candidates with moderate shrinkage
remain feasible on CUB. CIFAR rejects most candidates because at least one
fold regresses, while miniImageNet retains corrected-radius choices near
the covariance endpoint. Different parameter choices can produce similar accuracy,
so the selector need not identify a unique shrinkage coefficient. The supplement
provides target-oracle and leave-one-fold-out diagnostics. Reordering the novel
classes preserves the qualitative result, although these tests keep the
base--novel class partition fixed. The trainable controls in the supplement are sensitive to how examples are weighted.
Sample weighting gives the more numerous base examples greater influence.
Inverse-frequency weighting changes the balance between base and novel classes
and can reverse the overall ranking.

\subsection{Gains from unlabeled queries}
\label{sec:transductive-benchmark}

Table~\ref{tab:transductive-benchmark} reports each support-only baseline alongside
the additional gain from each query-assisted method.
With cosine NCM, QPR adds $0.67$, $0.37$, and $0.54$ session-average points
on CUB-200, CIFAR-100, and miniImageNet, respectively. With \ours{}, QPR improves
session-average accuracy by $1.63$ percentage points on CUB-200, $2.03$ on CIFAR-100,
and $1.21$ on miniImageNet. QPR improves both support-only heads, with larger gains
from \ours{} in these experiments.

The highest-performing method varies with the starting head and dataset. With
\ours{}, self-excluded QPR performs best on CUB, ordinary QPR on miniImageNet, and
TIM-GD on CIFAR-100. On CIFAR-100, TIM-GD reaches $76.37$, compared with $76.10$ for
QPR. With cosine NCM, TIM-GD also performs best on CIFAR-100 and miniImageNet.

Excluding each query from its own update gives positive gains with both heads
on all three datasets. On CUB, it also improves accuracy over ordinary QPR.
A query's inclusion in its own update
is therefore unnecessary for the observed benefit.

The transferred methods are commonly evaluated in balanced low-way
episodes. All-seen FSCIL instead combines many-shot base classes with a
growing few-shot novel block. The native five-way check confirms that
TIM-GD improves in that setting, but does not isolate why its all-seen
gains vary across heads and datasets. QPR targets the base--novel asymmetry by
updating only eligible novel directions. Those updates can also move a
novel boundary across base queries, producing a base-accuracy cost.

In the reported ablations, re-estimating radii from pseudo-labeled queries
reduces performance on all three datasets. Giving all queries within an admitted group equal weight, changes accuracy very little. Performance drops when all conservative controls are removed together. These ablations support keeping the selected geometry fixed, but they
do not establish that every individual gate is necessary on every dataset.

With the complete query cohort (Fig.~\ref{fig:qpr-operating-boundary}a),
CUB and CIFAR gains peak at two supports; miniImageNet is similar at one
and two. At $K=5$ (Fig.~\ref{fig:qpr-operating-boundary}b), very small
balanced cohorts yield little or no change, five-query cohorts can hurt,
and ten or twenty queries per class give positive gains. The joint grid
(Fig.~\ref{fig:qpr-shot-cohort-grid}) shows that larger cohorts help most
when supports are scarce. This pattern is consistent with less remaining
prototype error at larger $K$ and more reliable query assignments in
larger cohorts.

The CLIP backbone extensions show the same qualitative pattern. On DTD,
QPR improves its parent even though \ours{} itself underperforms LDA.

\subsection{Boundary and cost checks}
\label{sec:cost}

The base-only ResNet-18 control does not satisfy the majority criterion on
the standard datasets. Resource measurements use cached ViT-B/16 features.
\ours{} stores a pooled eigensystem and class prototypes, and adds novel
means in closed form. The supplement reports  cost of classifier after
feature extraction. This doesn't include end to end latency. Also QPR requires access to the query cohort.

\section{Limitations}
\label{sec:limitations}

We cannot verify whether CLIP's pretraining data exclude the novel classes or not. The additional encoder families also use ImageNet supervision. The
base-only ResNet excludes novel benchmark classes during training, but its features
are also weaker. This control therefore cannot separate the effect of pretraining
exposure from representation quality. The complete-data means and linear probe serve as retrospective references. The two accuracy changes add up exactly to the tested gap in the chosen order, but this accounting alone does not establish their causal contributions. \ours{}’s gains over LDA are sometimes small in practice, and its selector only approximates how competition changes as novel classes are added. Our transfer evaluation also uses a fixed split of class identities into base and novel groups. \qpr{} still requires an unlabeled query batch, even when each query is excluded from its own update, and adjusting novel prototypes can reduce accuracy on base classes.
\FloatBarrier
\section{Conclusion}
\label{sec:conclusion}

Across the strong frozen representations we tested, replacing novel-class means accounted for more of the accuracy gap than changing the classifier in our ordered comparisons. This finding motivates two focused corrections. \ours{} selects prototype radius and covariance geometry using labeled data, with an exact fallback to LDA. \qpr{} uses an unlabeled query cohort to refine novel-class prototype directions while preserving that geometry. Together, the results suggest that no single classifier is best for every setting. Effective frozen-feature FSCIL depends on using the available evidence to estimate classes with few labeled examples reliably.

\FloatBarrier
\section*{Code Availability}
The implementation and reproduction instructions are publicly available at
\url{https://github.com/Trex1102/fscil-prototype-estimation}.

{
    \small
    \bibliographystyle{ieeenat_fullname}
    \bibliography{main}
}

\end{document}